# Segmentation Regularized Training for Multi-Domain Deep Learning Registration applied to MR-Guided Prostate Cancer Radiotherapy

Sudharsan Madhavan, Chengcheng Gui, Lando Bosma*, Josiah Simeth, Jue Jiang, Nicolas Côté, Nima Hassan Rezaeian, Himanshu Nagar, Victoria Brennan, Neelam Tyagi[+], Harini Veeraraghavan[+]

Memorial Sloan Kettering Cancer Center, New York, NY, USA
&
*University Medical Center Utrecht (UMC Utrecht), Utrecht, The Netherlands
±
[+]Equal senior

## Highlights

- Segmentation regularized training of a deep learning deformable image registration (DL-DIR) improves generalizable applicability to MRI acquired on 1.5 Tesla and 3.0 Tesla imagers
- DL-DIR model was more accurate than variational registration methods, producing accuracy within 3 mm.
- DL-DIR model showed preliminary feasibility to perform deformable dose accumulation for patients with prostate cancer across 5 treatment fractions
- Dose accumulation results showed feasibility to evaluate compliance of delivered doses to institutional dose constraints

## Abstract

**Background and Purpose:** Accurate deformable image registration (DIR) is required for contour propagation and dose accumulation for magnetic resonance guided adaptive radiotherapy (MRgART). The goal of this study was to train and evaluate a deep learning (DL) DIR method using contour guided regularization for domain invariant MR-MR registration.

**Materials and Methods:** A progressively refined registration and segmentation (ProRSeg) DL-DIR method was trained with 262 pairs of 3 Tesla MR simulation (MR-Sim) scans of patients with prostate cancer. A weighted segmentation consistency loss was included for regularizing training. ProRSeg was tested on same- (58 pairs), cross- (72 1.5 Tesla MR Linac [MRL] pairs), and mixed-domain (42 MRSim-MRL pairs) datasets to evaluate contour propagation accuracy for clinical target volume (CTV), bladder, and rectum. Dose accumulation was performed for 42 patients undergoing 5-fraction MRgART.

**Results:** ProRSeg demonstrated generalization for the bladder, achieving statistically similar mean Dice Similarity Coefficients (DSCs) across same-, cross-, and mixed-domains (0.88, 0.87, and 0.86, respectively). For the rectum and CTV, performance was domain-dependent: accuracy was significantly higher on the cross-domain MRL dataset (DSCs 0.89) compared to the same-domain training data, while the mixed-domain task (MRSim-MRL) proved most challenging. The model's strong cross-domain performance prompted us to study the feasibility of using it for dose accumulation. ProRSeg-based dose accumulation indicated that 83.3% of patients met the CTV coverage constraint (D95 ≥ 40.0 Gy) and the bladder sparing constraint (D50 ≤ 20.0 Gy). Although all patients (100%) achieved the minimum mean target dose (Mean > 40.4 Gy), only 9.5% remained under the upper limit (< 42.0 Gy).

**Conclusions:** ProRSeg showed reasonable performance for multi-domain MR-MR registration for prostate cancer patients. Dose accumulation analysis indicated preliminary feasibility to evaluate compliance of delivered treatments to clinical constraints.

# Introduction

Magnetic Resonance Imaging (MRI) guided adaptive radiotherapy (MRgART) is a new treatment modality. Availability of an onboard MR imager with hybrid linear accelerator combined with superior soft-tissue contrast compared to conventional CT imaging allows for opportunistic daily adaptation of radiation treatments [1,2]. However, practical clinical implementation of daily adaptation requires fast and accurate segmentation of organs at risk (OARs). Deformable image registration (DIR) is a frequently used method to propagate contours from prior treatment fractions [3–5]. Moreover, DIR can be used to perform dose accumulation [6–8] that is useful to assess quality of delivered treatments as well as inform treatment adaptation.

Variational DIR as well as deep learning DIR (DL-DIR) methods have been developed for MR to MR registration applied to MRgART for pelvis and abdominal cancers [6,9,10]. Unlike variational methods, DL-DIR methods directly compute the transformation between a pair of images, obviating the need for iterative optimization of image alignment following training. Unsupervised DL-DIR methods that do not require "ground truth" deformation vector fields to extract models to compute a dense deformation between image pairs are a practical option. These methods perform training by maximizing image similarity between resampled and target images to optimize the networks [10–12]. However, MR images depict considerable variations in MR signal intensities, which can adversely impact accuracy of aforementioned DL-DIR methods when applied to datasets with different intensity characteristics [6,13].

Domain invariant registration can also be achieved by either training the model with datasets representative of the different imaging variations. Domain adaptation techniques employ generative adversarial networks to create variations in the training images to enhance generalizability [14,15]. However, such an approach requires additional networks and training data to transform between different domains.

A second set of approaches uses the geometry of the structures to constrain the registration process. The latter strategy has shown to be effective in both iterative registration methods [16–19] and deep learning frameworks [20–22]. For instance, including geometry or contour information explicitly as an input [16,23] or through segmentation driven losses [20] has been shown to improve registration accuracy. Hence, we adopted a joint registration-segmentation approach called ProRSeg previously shown to be effective for aligning MR-MR scans of patients with pancreatic cancers [10] for aligning pelvic MRI scans of patients with prostate cancers. Geometry regularization is applied as a loss to train the registration and segmentation network. The segmentation network trained in parallel provides additional regularization to the registration network as a multi-task objective. Following training, only the registration network is kept for testing.

Whereas prior works have typically focused on evaluating the accuracy gains of adding contour guidance with respect to other approaches, we evaluated the capability of the approach to be used without any further refinement on different MR acquisitions. For this purpose, ProRSeg was trained with longitudinal MR scans acquired on a diagnostic quality 3 Tesla MR simulator (MR-Sim). ProRSeg training was enhanced with a weighted segmentation regularization loss applied to registration propagated contours. The same model was then evaluated on a different or cross-domain MR-Linac (MRL) (1.5 Tesla - 1.5 Tesla), and mixed-domain (3 Tesla MR-Sim - 1.5 Tesla MRL) datasets. In addition, we evaluated the feasibility to perform deformable dose accumulation and compared accumulated doses to institutional constraints as a preliminary plan quality analysis.

# Methods and Materials

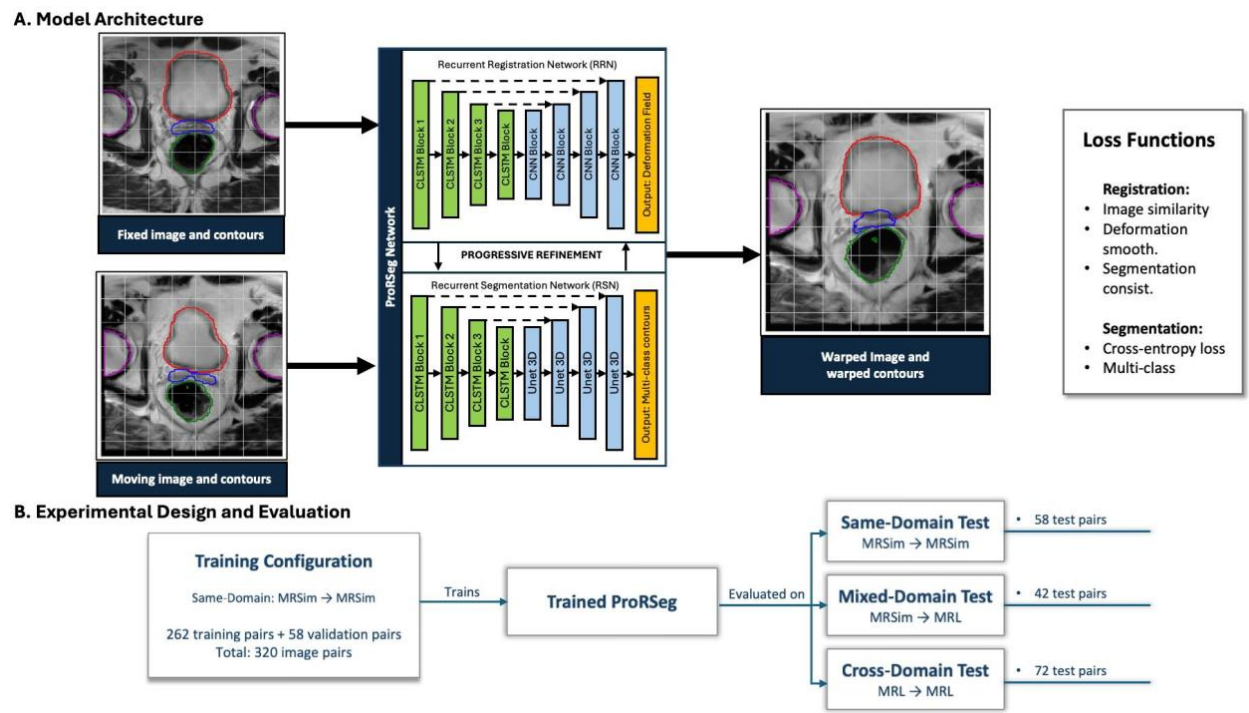

**Figure 1a.** Flow diagram illustrating dataset partitioning for registration evaluation for same-domain, mixed-domain, and cross-domain analysis. **b.** Schematic representation of Progressively Refined joint registration segmentation (ProRSeg) architecture applied to multi-domain MR-guided prostate radiotherapy. The framework consists of recurrent registration (RRN) and segmentation (RSN) networks to simultaneously register and segment organs at risk.

## 2.1. Image Acquisition, Dataset Preparation, and Experimental framework

Discovery dataset consisted of 3 Tesla 2D T2-weighted (T2WI) MRI (TR: 4740 ms, TE: 100 ms) acquired for treatment planning simulation (MRSim). This dataset consisted of 320 image volume pairs from 34 patients with prostate cancers acquired at treatment planning simulation and longitudinally every 3 months following radiation treatment up to a maximum of 24 months.

A portion of data was set aside for training (262 image pairs) and held-out testing (58 image pairs) for same-domain testing as shown in Figure 1b. Cross domain testing was performed using 3D T2WI MRI acquired on 1.5 Tesla MR-Linac (MRL) images (TR: 1300 ms, TE: 87 ms). This dataset consisted of 72 pairs from patients imaged every day on a 5 fraction treatment. Mixed domain testing was performed on a subset of 42 patients by aligning MRSim to the first treatment fraction MRI from MRL (Figure 1b).

For the dose accumulation feasibility study, a separate cohort of 42 patients undergoing a 5-fraction hypofractionated MRgART course was analyzed. All daily MRL scans and daily online adaptive treatment plans for these patients were used to evaluate the dose mapping and accumulation methodology. Our daily online adaptive workflow was described in detail in our prior publication [24].

### 2.1.1. Preprocessing Pipeline

All images underwent a standardized preprocessing workflow designed to ensure dimensional consistency and intensity normalization across different scanning platforms. The preprocessing steps included:

1. **Resampling**: Images and corresponding segmentation masks were resampled to uniform dimensions of 256×256×128 pixels. Trilinear interpolation was used for image data and nearest-neighbor interpolation was used for segmentation masks to preserve discrete label values. Spacing adjustments were calculated using the original affine transformation matrices to ensure anatomical consistency.

2. **Intensity Normalization**: A percentile-based normalization approach was employed by clipping intensities below 10th and 90th percentiles of MR signal intensities and normalizing the resultant image to be within a standardized range [0,1].

## 2.2. ProRSeg network architecture and training details

### 2.2.1. Network Architecture

Figure 1a shows the ProRSeg architecture implemented to align pelvic MRI and used for multi-domain evaluation applied to prostate radiotherapy with our multi-domain training strategy. The recurrent registration network (RRN) computes deformation vector fields through 8 CLSTM iterations, while the recurrent segmentation network (RSN) trained in parallel with the RRN generates organ segmentations through 9 CLSTM iterations. Both networks are optimized simultaneously. The RRN was optimized using image similarity ($L_{sim}$), deformation smoothness ($L_{smooth}$), and weighted segmentation consistency ($L_{cons}$).

The smoothness loss is implemented to ensure smooth deformation vector fields and is computed as:

$$L_{\text{smooth}} = \sum_{p \in \Omega} \|\nabla \phi(p)\|^2.$$

Image similarity loss computes the mean square error between image patches. Specifically, the image is divided into non-overlapping image patches and the similarity of patches from the resampled and target image are computed.

The segmentation consistency loss in ProRSeg assumed equal contribution for all analyzed structures and computed the error as multi-category cross entropy loss. The current work weighted the contribution of different structures in order to give higher importance to organs such as the bladder undergoing larger deformation than the rectum and the prostate clinical target volume (CTV). The segmentation consistency loss was then computed as a weighted DSC overlap loss that compared the registration propagated segmentation and the target image segmentation. Each CLSTM step is connected to a spatial transform network to generate an incremental deformation vector field, which are used to generate progressively refined propagated contours. Hence, the segmentation consistency loss was computed from the individual segmentations, thus providing a deep supervision segmentation consistency loss. The segmentation consistency loss is thus computed as:

$$L_{\text{cons}} = \sum_{i=0}^{N} \sum_{k \in \mathcal{K}} w_k \left(1 - \text{DSC}\left(y_{f,k}, g\left(x_m^i, y_{m,k}^i, x_m, h_g^i\right)\right)\right),$$

Where:

$\mathcal{K} = \{\text{bladder}, \text{rectum}, \text{prostate}\}$,

$w_{\text{bladder}} = 0.40, \quad w_{\text{rectum}} = 0.30, \quad w_{\text{prostate}} = 0.30.$

Here DSC represents the Dice similarity coefficient, $(y_f)$ is the target or fixed image segmentation, $(g())$ is the RRN, $(x_m^i)$ and $(y_m^i)$ are the moving image and the propagated contour produced after CLSTM step $(i)$, using the hidden state $(h_g^i)$ corresponding to the CLSTM step i.

The total registration loss combines image similarity, deformation smoothness, and segmentation consistency:

$$L_{\text{total}} = L_{\text{sim}} + \lambda_{\text{smooth}} L_{\text{smooth}} + \lambda_{\text{cons}} L_{\text{cons}},$$

The segmentation network is optimized with multi-category cross-entropy loss at each CLSTM step (*i*), providing deep supervision throughout the progressive refinement process by focusing computational resources on the most challenging voxels:

$$L_{\text{seg}} = \sum_{t=0}^{N} \log P\left(y_f | s\left(x_t^i, y_t^i, x_m, h_m^i\right)\right),$$

where $s()$ represents the segmentation function.

### 2.2.2. Hyperparameter Optimization

Hyperparameter optimization was performed using grid search to appropriately weight the smoothness (λ_smooth) and segmentation consistency (λ_cons) losses by evaluating a range of weights from 10 to 30 for λ_smooth and 1 to 10 for λ_cons. The best model selected from the validation set resulted in a λ_smooth of 30 and λ_cons of 5.

## 2.3 Registration Assessment and Analysis Methods

Testing was performed on a held-out set of same-domain, cross domain and mixed domain datasets as described in Section 2.1 and as shown in Figure 1b. ProRSeg was evaluated on the same-domain dataset against rigid (as a baseline reference), as well as two established iterative registration methods: SyN available through the open-source ANTsX [25,26] and EVolution or EVO [27] that use intensity based metrics to perform deformable alignment.

### 2.3.1. Geometric metrics

Commonly used geometric overlap metrics including the Dice similarity coefficient (DSC), 95th Percentile Hausdorff Distance (HD95), and Mean Distance to Agreement (MDA) were calculated for various OARs including bladder, rectum, and the CTV. As a secondary analysis, accuracy of propagating the contours for urethra was also computed.

### 2.3.2 Dose Mapping and Accumulation Methodology

Dose mapping was performed by utilizing the deformation vector fields (DVFs) generated by the model. The process followed multiple sequential steps: First, voxel-wise mapping was performed where, for each patient, the DVFs computed between different fraction images established spatial correspondence between anatomical points across treatment fractions. Next, each fraction's dose distributions were interpolated to the reference anatomy (fraction 1) using trilinear interpolation. Finally, after mapping the doses to the same coordinate system, the dose distributions from individual fractions were summed to generate the accumulated dose distribution, accounting for the entire treatment course. The dosimetric parameters were quantified using mean and standard deviation.

### 2.3.3 Statistical Analysis

Statistical comparisons were performed using two-sided, paired Wilcoxon signed rank tests with statistical significance defined at p < 0.05 to measure the difference between DSC and HD95 metrics produced for bladder, rectum, and CTV using the different methods. To compare performance across the same-, cross-, and mixed-domain datasets, statistical significance was determined using the two-sided, unpaired Wilcoxon rank-sum test, with p < 0.05 considered significant.

# Results

## 3.1. Quantitative Comparison of contour alignment

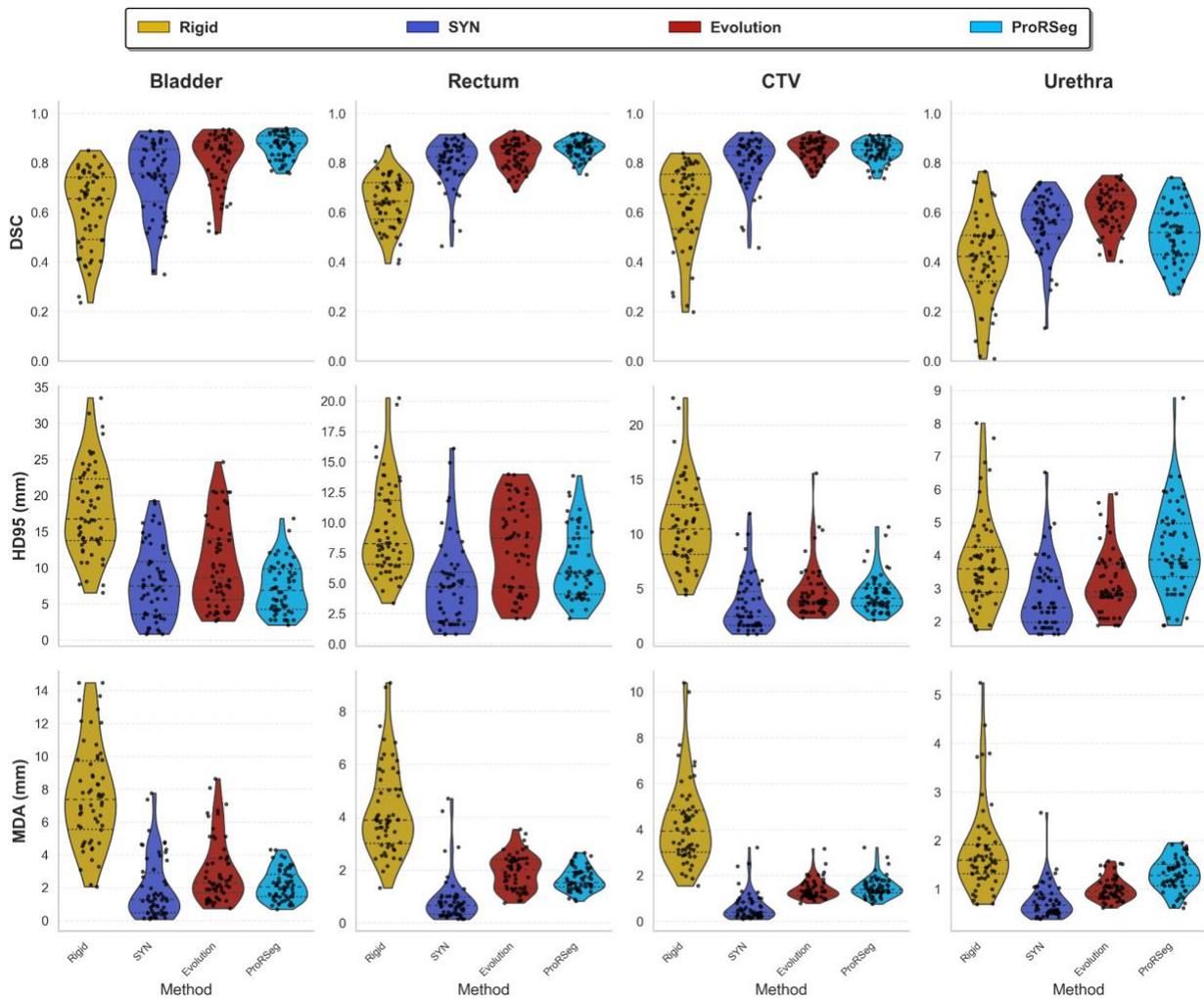

**Figure 2. Quantitative comparison of contour alignment produced using the DIR methods for multiple pelvic organs using DSC (top row), HD95 (middle row) and MDA (bottom row).**

Figure 2 shows quantitative comparison of ProRSeg against the baseline and iterative methods on the same-domain (MRSim-MRSim) test set. Registration methods (excluding Rigid) were similarly accurate for the CTV exceeding a mean DSC of 0.80, with ProRSeg achieving a mean DSC of 0.85 ± 0.04. Differences in accuracies were more apparent for organs depicting large anatomical differences including the bladder (ProRSeg: DSC of 0.88, HD95 of 2.99 mm; Rigid: DSC 0.61, p < 0.001, HD95 17.94 mm, p < 0.001; SYN: DSC 0.74, p < 0.001, HD95 7.97 mm, p = 0.273; EVolution: DSC 0.81, p < 0.001, HD95 10.29 mm, p < 0.001) and the rectum (ProRSeg: DSC of 0.86, HD95 of 2.50 mm; Rigid: DSC 0.64, p < 0.001, HD95 9.35 mm, p < 0.001; SYN: DSC 0.80, p < 0.001, HD95 4.84 mm, p < 0.001; EVolution: DSC 0.83, p < 0.001, HD95 8.00 mm, p < 0.01). ProRSeg was significantly more accurate than all other methods for bladder and rectum, indicating superior performance in handling large anatomic variability. ProRseg produced slightly more accurate segmentation for the urethra (DSC 0.52, HD95 3.05 mm) compared to other methods. However, the performance was not statistically different indicating variable performance for all methods.

### 3.2. Qualitative Assessment of Registration Performance

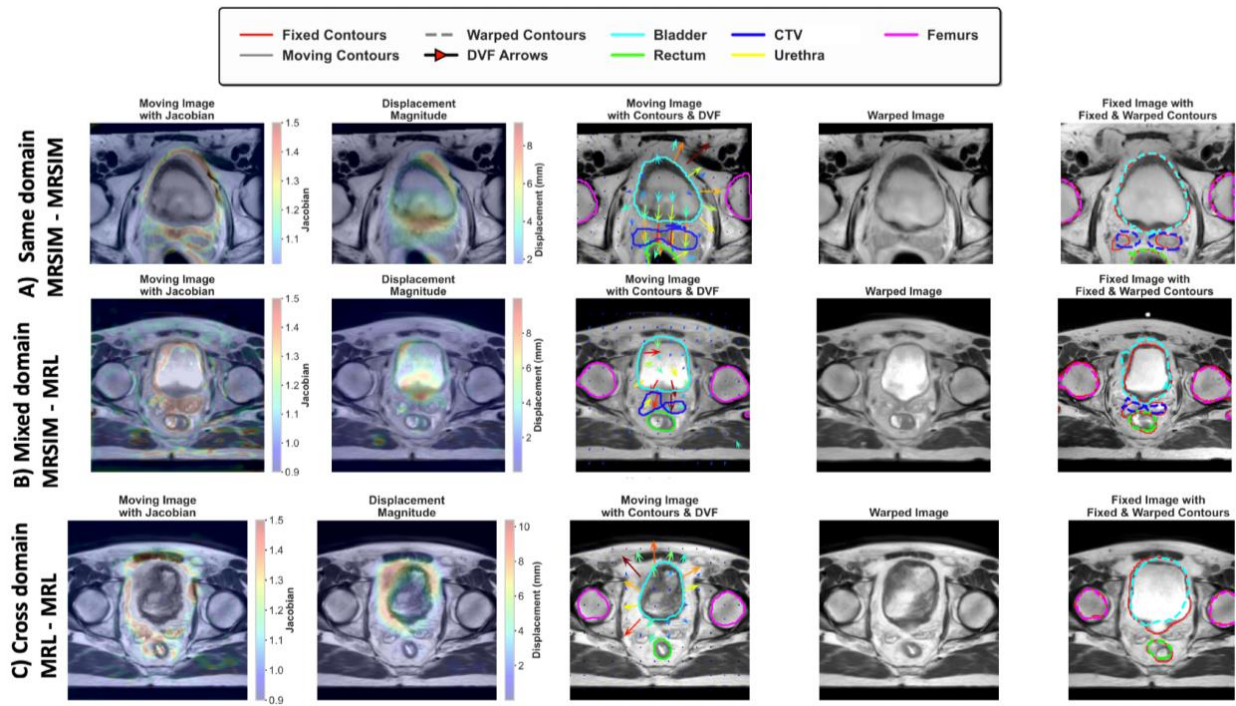

**Figure 3. Representative cases showing results of ProRSeg together with Jacobian determinant maps and displacement maps for same-domain (top), cross domain (middle), and mixed domain (bottom row) datasets.**

Representative examples of registration obtained using ProRSeg for same-domain MRSim-MRSim Figure 3(A), mixed-domain MRL-MRL Figure 3(B), and cross-domain MRSim-MRL Figure 3(C) show the moving image with the displacement vectors, deformed contours, Jacobian determinant and displacement magnitude maps. The Jacobian determinant maps show large deformations occurring in the vicinity of organs like the bladder and rectum with values ranging from 0.9 to 1.5, and displacement magnitude maps that show large displacements (magnitudes ranging from 2 mm to 10 mm) occurred within the bladder and at the boundary of the rectum for all three cases.

## 3.3. Generalization accuracy for same, cross and mixed domain datasets

Table 1. Accuracy of ProRSeg based contour alignment applied to three different domains. *p < 0.05 compared to Same-domain; †p < 0.05 compared to Mixed-domain (two-sided, unpaired Wilcoxon rank-sum test).

| Organ | Same domain (MRSim-MRSim) | | | Mixed domain (MRSim-MRL) | | | Cross domain (MRL-MRL) | | |
|---|---|---|---|---|---|---|---|---|---|
| | DSC | HD95 | MDA | DSC | HD95 | MDA | DSC | HD95 | MDA |
| Bladder | 0.88 ± 0.05 | 2.99 ± 1.88 | 0.46 ± 0.34 | 0.86 ± 0.07 | 2.58 ± 1.94 | 0.44 ± 0.39 | 0.87 ± 0.06 | 2.39 ± 2.08 | 0.37 ± 0.42 |
| Rectum | 0.86 ± 0.04 | 2.50 ± 1.55 | 0.36 ± 0.20 | 0.78 ± 0.05* | 3.35 ± 1.15* | 0.61 ± 0.21* | 0.89 ± 0.03*† | 1.42 ± 1.04*† | 0.21 ± 0.15*† |
| CTV | 0.85 ± 0.04 | 1.95 ± 0.88 | 0.31 ± 0.19 | 0.82 ± 0.07 | 2.12 ± 0.79 | 0.36 ± 0.20 | 0.89 ± 0.05*† | 1.29 ± 0.72*† | 0.19 ± 0.14*† |
| Urethra | 0.51 ± 0.11 | 3.97 ± 1.82 | 1.25 ± 0.73 | 0.39 ± 0.16* | 4.00 ± 2.02* | 1.30 ± 0.59* | 0.64 ± 0.16*† | 1.85 ± 0.91*† | 0.49 ± 0.32*† |

The generalization accuracy of ProRSeg across the three domains is shown in Table 1. For the bladder, performance was statistically invariant, with high mean DSC scores observed across the same-, mixed-, and cross-domain datasets (0.88, 0.86, and 0.87, respectively; p>0.05 for all comparisons).

In contrast, the rectum, CTV, and urethra exhibited domain-dependent performance. Notably, registration accuracy in the cross-domain (MRL-MRL) setting for the rectum and CTV (mean DSC 0.89 for both) was higher than in the same-domain (p<0.001). The mixed-domain (MRSim-MRL) setting yielded the statistically significant lowest accuracy for these structures, with performance being lower than both other domains (p<0.05 for all relevant comparisons).

Finally, the accuracy of ProRSeg in the cross-domain (MRL-MRL) setting was benchmarked against a model instance created using MRL-MRL pairs (training = 288, validation = 72). Testing was performed on identical cases that showed statistically similar accuracies for (CTV: DSC of 0.88 ± 0.06 ; and rectum: DSC of 0.88 ± 0.05, p = 0.17). However, a difference was observed for the bladder, where the benchmark model achieved a statistically significant higher mean DSC (0.91 vs. 0.87, p<0.001).

## 3.4. Dosimetric Impact Assessment

### 3.4.1 Dose-Volume Histogram Analysis

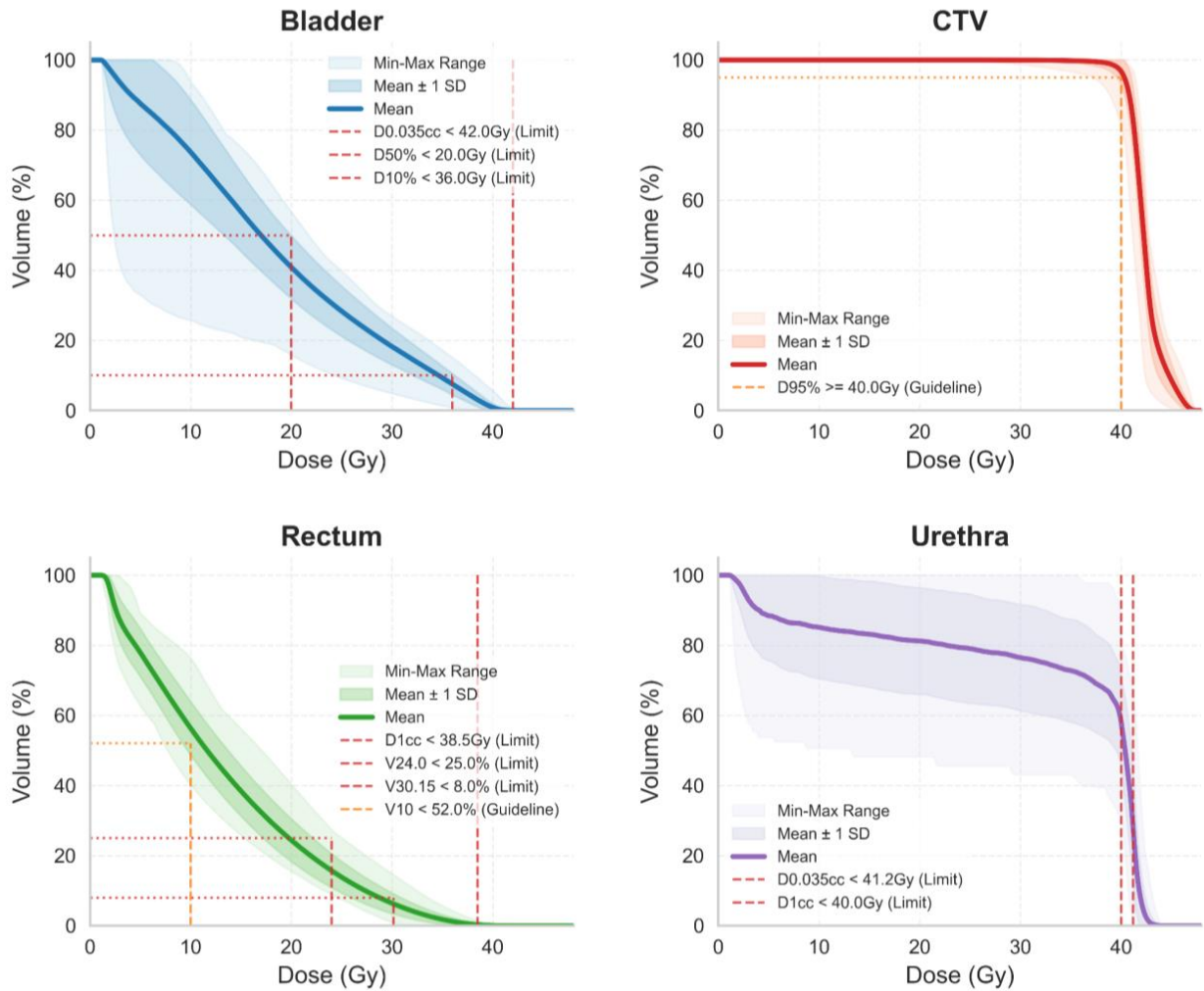

Figure 4. Dose-volume histograms (DVHs) derived using dose accumulation performed with ProRSeg for bladder, CTV, rectum, and urethra. Mean (solid lines), ± one standard deviation (light shaded areas), and min-max range (lightest shaded regions) are shown.

Dose-volume histogram analysis derived from ProRSeg-based deformable registration revealed critical insights into organ-specific dose distributions (Figure 4). The bladder exhibited a gradual dose gradient with substantial inter-patient variability, with mean values indicating approximately 40% of bladder volume receiving 20 Gy. This heterogeneity reflects the anatomical variations successfully captured by the segmentation-regularized registration algorithm.

For the clinical target volume prostate (CTV), near-uniform dose coverage was achieved across the patient cohort, with the mean curve indicating that 95% of the target volume consistently received at least

40.3 Gy. The narrow standard deviation range suggests the adequate target coverage achieved through MR-guided radiotherapy planning, despite variations in patient anatomy in our dataset.

Analysis of rectum DVHs revealed generally favorable sparing, with mean values demonstrating that only 25% of rectal volume received doses exceeding 20 Gy. Notably, all patients satisfied the critical V30 constraint, while the majority met the V24 constraint (volume receiving 24 Gy). The relatively wider standard deviation range in the mid-dose region (10-30 Gy) highlights the importance of patient-specific adaptation strategies during treatment.

Urethra dose distributions displayed a characteristic pattern reflecting its anatomical position within the high-dose region, with approximately 80% of urethral volume receiving 30 Gy before exhibiting a sharp fall-off after 40 Gy.

### 3.4.4 Dose Constraint Compliance Analysis

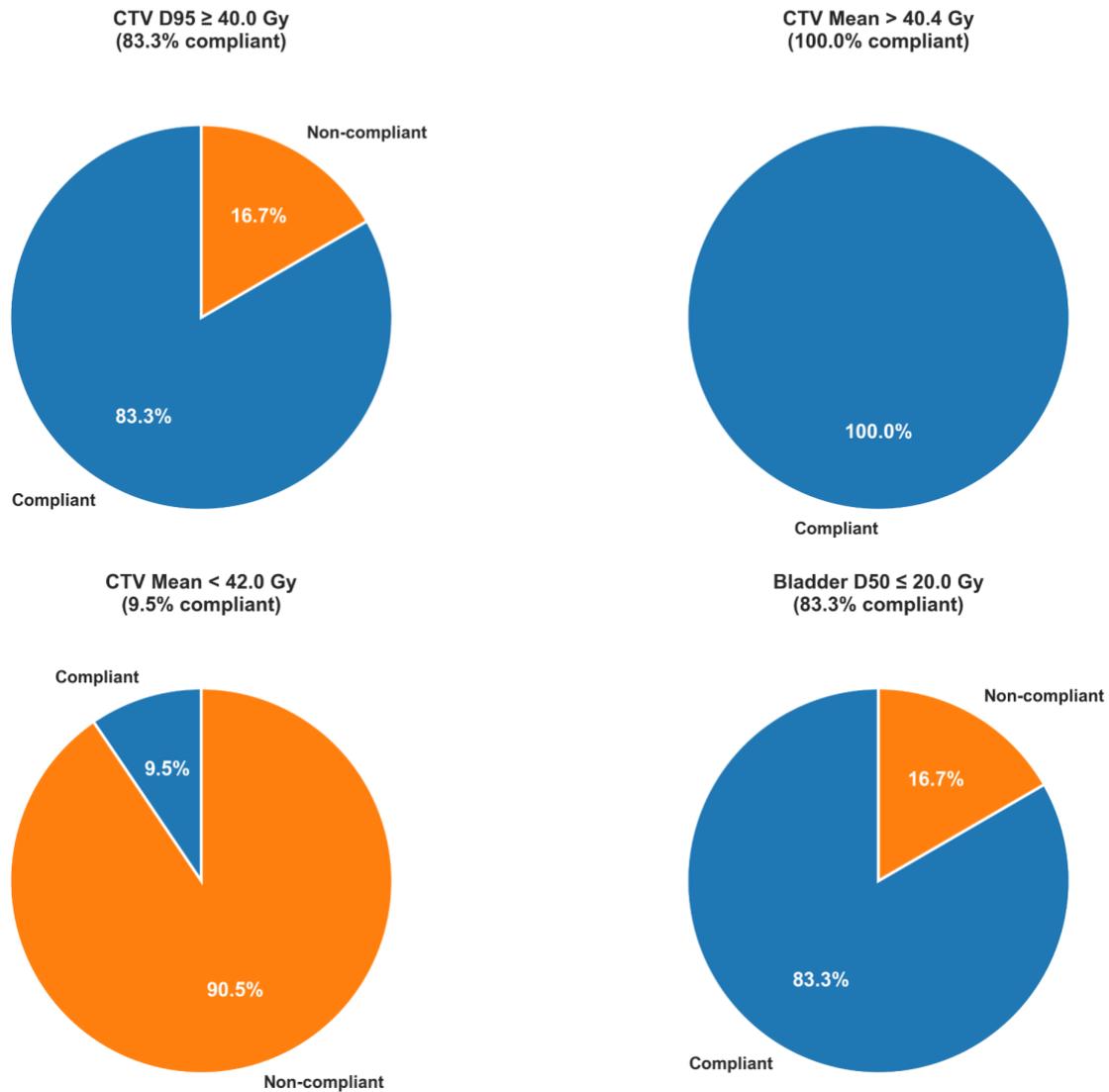

**Figure 5. Compliance rates for institutional dose constraints based on the accumulated dose distributions (n = 42 patients) using ProRSeg for CTV and bladder.**

Assessment of institutional dose constraint compliance following deformable dose accumulation revealed important insights regarding treatment quality and potential areas for optimization (Figure 5). Specifically, our analysis showed a high compliance for the mean target dose, with 100% of cases achieving the standard mean target dose (CTV Mean ≥ 40.4 Gy). The target coverage requirement (CTV D95 ≥ 40.0 Gy) was achieved in 83.3% of cases. A more stringent evaluation criterion (CTV Mean < 42.0 Gy) was achieved in only 9.5% of cases, indicating that while adequate coverage was maintained, dose escalation beyond institutional standards was uncommon with current treatment planning protocols. This observation identifies a potential pathway for future protocol enhancement should clinical outcomes suggest benefits from dose intensification. Regarding organs at risk, the bladder dose constraint (D50 ≤ 20.0 Gy) was satisfied in 83.3% of patients, revealing that a portion of patients experienced higher than desired bladder exposure despite adaptive planning. This finding highlights the need for continued refinement of bladder-sparing techniques in specific anatomical scenarios.

This detailed dosimetric assessment was performed on the cross-domain (MRL-MRL) dataset, for which the registration model demonstrated high geometric accuracy. The resulting dose accumulation, which accounts for interfraction anatomical variations, provides valuable feedback for treatment protocol refinement and identifies specific subgroups that might benefit from modified planning strategies.

## Discussion

This study evaluated the multi-domain generalization of a segmentation-regularized DIR model, ProRSeg, for MR-guided prostate radiotherapy. Although our model achieved good accuracy across all domains, multi-domain generalization performance was organ dependent. The statistically invariant performance for the bladder suggests that the learned anatomical features could be robust for organs with clear and easy boundaries, regardless of the MR imaging platform. A key and counterintuitive finding was that the model's accuracy for the rectum and CTV was slightly higher in the cross-domain (MRL-MRL) setting than in the same-domain (MRSim-MRSim) setting it was trained on. This suggests that the features learned from 3T MR-Sim data may be useful when applied to the 1.5T MRL images, which perhaps present a more consistent registration task with less anatomical variability between daily fractions compared to the long-term follow-up scans in the same-domain dataset. Conversely, the mixed-domain (MR-Sim to MRL) setting was identified as the primary performance bottleneck, consistently yielding the lowest accuracy for the rectum, CTV, and urethra. This highlights that while the model generalizes well within scanner types, registering directly across different field strengths and acquisition protocols remains a significant challenge. This finding underscores the importance of dedicated domain adaptation strategies to bridge this performance gap, representing a clear avenue for future work.

Our findings align with recent advances in segmentation-guided registration approaches. [14] demonstrated the effectiveness of using segmentation information with transfer learning for registration tasks, while [28] showed that structure guidance constraints improve registration accuracy in adaptive radiotherapy settings. The segmentation consistency loss employed in our work builds upon these concepts by enforcing anatomical plausibility during the registration process. Furthermore, the cross-domain generalization achieved in our study addresses a critical challenge identified in recent domain adaptation surveys for medical imaging [29], where maintaining performance across different imaging protocols remains a significant barrier to clinical deployment.

The organ-specific weighting scheme (bladder: 0.40, rectum: 0.30, prostate: 0.30) is particularly important for structures with varying anatomical complexity and clinical importance. The higher weight assigned to the bladder reflects its greater anatomical variability between fractions, while equal weights for rectum and prostate balance their clinical significance in dose optimization.

Beyond cross-domain generalization, ProRSeg showed more accurate performance compared to multiple intensity based iterative registration methods that did not use contour guidance. However, all methods demonstrated acceptable accuracy for the prostate CTV. ProRSeg achieved significantly better alignment of the bladder and rectum, which exhibit greater anatomic variability between treatment fractions. ProRSeg showed feasibility to derive dose metrics for detecting clinically relevant dosimetric variations. Wide variations were observed for bladder and the urethra. Our approach identified high compliance for the mean CTV dose target (100%), while the CTV D95 coverage and bladder sparing constraints were met in 83.3.% of cases. Such quality analyses could be useful for assessing accuracy of delivered treatments and for integrating accumulated dose information for adaptive treatments.

Our study has a few limitations. First, the analysis of domain generalization only considered the impact of MR images acquired on two different platforms with different magnet strengths but not the wider set of variations in MRI. The dose accumulation and compliance analysis are only a preliminary feasibility study performed to assess applicability of DL-DIR method. Further analysis on larger cohorts as well as multi-institutional cohorts is needed to demonstrate efficacy. Our study demonstrated robust cross - domain performance to align MR scans across different MR platforms used in prostate cancer radiotherapy.

# Conclusion

In conclusion, our segmentation-regularized registration approach demonstrates robust performance across different MR imaging platforms used in radiotherapy, addressing a critical challenge in the clinical implementation of MR-guided adaptive workflows. The integration of anatomical segmentation as a regularization mechanism significantly enhances cross-domain generalization, enabling accurate registration despite variations in image characteristics. These findings support the feasibility of using deep learning-based registration for organ tracking and dose accumulation across different MRI platforms in radiotherapy, potentially improving treatment precision and clinical outcomes.

# References

https://doi.org/10.1016/j.media.2021.102041
[1]   Tetar SU, Bruynzeel AME, Lagerwaard FJ, Slotman BJ, Bohoudi O, Palacios MA. Clinical implementation of magnetic resonance imaging guided adaptive radiotherapy for localized prostate cancer. Physics and Imaging in Radiation Oncology 2019;9:69–76. https://doi.org/10.1016/j.phro.2019.02.002.

[2] Kishan AU, Ma TM, Lamb JM, Casado M, Wilhalme H, Low DA, et al. Magnetic Resonance Imaging–Guided vs Computed Tomography–Guided Stereotactic Body Radiotherapy for Prostate Cancer: The MIRAGE Randomized Clinical Trial. JAMA Oncology 2023;9:365–73. https://doi.org/10.1001/jamaoncol.2022.6558.
[3] Brock KK, Mutic S, McNutt TR, Li H, Kessler ML. Use of image registration and fusion algorithms and techniques in radiotherapy: Report of the AAPM Radiation Therapy Committee Task Group No. 132. Medical Physics 2017;44:e43–76. https://doi.org/10.1002/mp.12256.
[4] Bohoudi O, Lagerwaard FJ, Bruynzeel AME, Niebuhr NI, Johnen W, Senan S, et al. End-to-end empirical validation of dose accumulation in MRI-guided adaptive radiotherapy for prostate cancer using an anthropomorphic deformable pelvis phantom. Radiotherapy and Oncology 2019;141:200–7. https://doi.org/10.1016/j.radonc.2019.09.014.
[5] Mittauer KE, Hill PM, Bassetti MF, Bayouth JE. Validation of an MR-guided online adaptive radiotherapy (MRgoART) program: Deformation accuracy in a heterogeneous, deformable, anthropomorphic phantom. Radiotherapy and Oncology 2020;146:97–109. https://doi.org/10.1016/j.radonc.2020.02.012.
[6] Murr M, Bernchou U, Bubula-Rehm E, Ruschin M, Sadeghi P, Voet P, et al. A multi-institutional comparison of retrospective deformable dose accumulation for online adaptive magnetic resonance-guided radiotherapy. Physics and Imaging in Radiation Oncology 2024;30:100588. https://doi.org/10.1016/j.phro.2024.100588.
[7] Leiner L, Fakultät HU (Germany) M. Dose accumulation of adapted treatment plans in MR-guided radiotherapy. Dose accumulation of adapted treatment plans in MR-guided radiotherapy, 2024.
[8] Malkov VN, Mansour IR, Kong V, Li W, Dang J, Sadeghi P, et al. Geometric and Dosimetric Validation of Deformable Image Registration for Prostate MR-guided Adaptive Radiotherapy 2025. https://doi.org/10.48550/arXiv.2504.07933.
[9] Zhang Y, Paulson E, Lim S, Hall WA, Ahunbay E, Mickevicius NJ, et al. A Patient-Specific Autosegmentation Strategy Using Multi-Input Deformable Image Registration for Magnetic Resonance Imaging–Guided Online Adaptive Radiation Therapy: A Feasibility Study. Advances in Radiation Oncology 2020;5:1350–8. https://doi.org/10.1016/j.adro.2020.04.027.
[10] Jiang J, Hong J, Tringale K, Reyngold M, Crane C, Tyagi N, et al. Progressively refined deep joint registration segmentation (ProRSeg) of gastrointestinal organs at risk: Application to MRI and cone-beam CT. Medical Physics 2023;50:4758–74. https://doi.org/10.1002/mp.16527.
[11] Balakrishnan G, Zhao A, Sabuncu MR, Guttag J, Dalca AV. VoxelMorph: A Learning Framework for Deformable Medical Image Registration. IEEE Trans Med Imaging 2019;38:1788–800. https://doi.org/10.1109/TMI.2019.2897538.
[12] de Vos BD, Berendsen FF, Viergever MA, Sokooti H, Staring M, Išgum I. A deep learning framework for unsupervised affine and deformable image registration. Medical Image Analysis 2019;52:128–43. https://doi.org/10.1016/j.media.2018.11.010.
[13] Zhong H, Kainz KK, Paulson ES. Evaluation and mitigation of deformable image registration uncertainties for MRI-guided adaptive radiotherapy. Journal of Applied Clinical Medical Physics 2024;25:e14358. https://doi.org/10.1002/acm2.14358.
[14] Mahapatra D, Ge Z. Training Data Independent Image Registration with Gans Using Transfer Learning and Segmentation Information. 2019 IEEE 16th International Symposium on Biomedical Imaging (ISBI 2019), 2019, p. 709–13. https://doi.org/10.1109/ISBI.2019.8759247.
[15] Ferrante E, Oktay O, Glocker B, Milone DH. On the Adaptability of Unsupervised CNN-Based Deformable Image Registration to Unseen Image Domains. In: Shi Y, Suk H-I, Liu M, editors. Machine Learning in Medical Imaging, Cham: Springer International Publishing; 2018, p. 294–302. https://doi.org/10.1007/978-3-030-00919-9_34.
[16] Bosma LS, Ries M, Denis de Senneville B, Raaymakers BW, Zachiu C. Integration of operator-validated contours in deformable image registration for dose accumulation in radiotherapy. Physics and Imaging in Radiation Oncology 2023;27:100483. https://doi.org/10.1016/j.phro.2023.100483.
[17] Zachiu C, Denis de Senneville B, Willigenburg T, Voort van Zyp JRN, de Boer JCJ, Raaymakers

BW, et al. Anatomically-adaptive multi-modal image registration for image-guided external-beam radiotherapy. Phys Med Biol 2020;65:215028. https://doi.org/10.1088/1361-6560/abad7d.
[18] Gu X, Dong B, Wang J, Yordy J, Mell L, Jia X, et al. A contour-guided deformable image registration algorithm for adaptive radiotherapy. Phys Med Biol 2013;58:1889. https://doi.org/10.1088/0031-9155/58/6/1889.
[19] Eichner T, Mörth E, Wagner-Larsen K, Lura N, Haldorsen I, Gröller E, et al. MuSIC: Multi-Sequential Interactive Co-Registration for Cancer Imaging Data based on Segmentation Masks. Eurographics Workshop on Visual Computing for Biology and Medicine 2022:81–91. https://doi.org/10.2312/VCBM.20221190.
[20] Elmahdy MS, Beljaards L, Yousefi S, Sokooti H, Verbeek F, Van Der Heide UA, et al. Joint Registration and Segmentation via Multi-Task Learning for Adaptive Radiotherapy of Prostate Cancer. IEEE Access 2021;9:95551–68. https://doi.org/10.1109/ACCESS.2021.3091011.
[21] Li Y, Fu Y, Gayo IJMB, Yang Q, Min Z, Saeed SU, et al. Semi-weakly-supervised neural network training for medical image registration 2024. https://doi.org/10.48550/arXiv.2402.10728.
[22] Gao Y, Sandhu R, Fichtinger G, Tannenbaum AR. A Coupled Global Registration and Segmentation Framework With Application to Magnetic Resonance Prostate Imagery. IEEE Transactions on Medical Imaging 2010;29:1781–94. https://doi.org/10.1109/TMI.2010.2052065.
[23] Hemon C, Rigaud B, Barateau A, Tilquin F, Noblet V, Sarrut D, et al. Contour-guided deep learning based deformable image registration for dose monitoring during CBCT-guided radiotherapy of prostate cancer. Journal of Applied Clinical Medical Physics 2023;24:e13991. https://doi.org/10.1002/acm2.13991.
[24] Brennan VS, Burleson S, Kostrzewa C, Godoy Scripes P, Subashi E, Zhang Z, et al. SBRT focal dose intensification using an MR-Linac adaptive planning for intermediate-risk prostate cancer: An analysis of the dosimetric impact of intra-fractional organ changes. Radiotherapy and Oncology 2023;179:109441. https://doi.org/10.1016/j.radonc.2022.109441.
[25] Tustison NJ, Cook PA, Holbrook AJ, Johnson HJ, Muschelli J, Devenyi GA, et al. The ANTsX ecosystem for quantitative biological and medical imaging. Sci Rep 2021;11:9068. https://doi.org/10.1038/s41598-021-87564-6.
[26] Avants BB, Epstein CL, Grossman M, Gee JC. Symmetric diffeomorphic image registration with cross-correlation: Evaluating automated labeling of elderly and neurodegenerative brain. Medical Image Analysis 2008;12:26–41. https://doi.org/10.1016/j.media.2007.06.004.
[27] Denis de Senneville B, Zachiu C, Ries M, Moonen C. EVolution: an edge-based variational method for non-rigid multi-modal image registration. Phys Med Biol 2016;61:7377. https://doi.org/10.1088/0031-9155/61/20/7377.
[28] Kuckertz S, Papenberg N, Honegger J, Morgas T, Haas B, Heldmann S. Learning Deformable Image Registration with Structure Guidance Constraints for Adaptive Radiotherapy. In: Špiclin Ž, McClelland J, Kybic J, Goksel O, editors. Biomedical Image Registration, Cham: Springer International Publishing; 2020, p. 44–53. https://doi.org/10.1007/978-3-030-50120-4_5.
[29] Guan H, Liu M. Domain Adaptation for Medical Image Analysis: A Survey. IEEE Transactions on Biomedical Engineering 2022;69:1173–85. https://doi.org/10.1109/TBME.2021.3117407.